\documentclass[preprint]{JASA}
\usepackage{tipa}
\usepackage{amsmath,amssymb,amsfonts}

\graphicspath{{figures/}}

\begin{document}

\title[Low-dimensional dynamics of tongue shape control]{A low-dimensional dynamical basis for tongue shape control}

\author{Sam Kirkham$^1$}
\email{s.kirkham@lancaster.ac.uk}
\author{Patrycja Strycharczuk$^2$}

\affiliation{
    $^1$Phonetics Laboratory, Lancaster University, United Kingdom \\
    $^2$Linguistics and English Language, University of Manchester, United Kingdom
    }

\date{\today}

\begin{abstract}
Speech production involves the control of a high-dimensional articulatory system with many degrees-of-freedom, but a widespread hypothesis is that speakers navigate this complexity using a set of lower-dimensional control variables. Common candidates for these control variables include individual articulators (e.g.~tongue body) and abstract tract variables (e.g.~tongue body constriction location), but identifying the most appropriate coordinate system for articulatory control remains an open question. In this study, we advance a model in which holistic tongue shape modes represent a potential low-dimensional control basis for a simple dynamical law. We first discover empirical tongue shape modes in a cross-speaker coordinate space and then fit the model to ultrasound data on monophthongal vowels from 36 speakers of British English. A three-dimensional linear point attractor fitted to these time-varying modes recovers the dynamics of tongue shape with a high degree of accuracy. This suggests that complex tongue movement can arise from independent modes that linearly combine under simple dynamical laws, thereby relocating some complexity from the control signal to the coordinate system. We show that this framing explains why one-dimensional articulatory measures can exhibit observable nonlinearities, and hypothesize that shape modes are a plausible kinematic expression of underlying muscle synergies.
\end{abstract}

\maketitle

\section{Introduction}

The production of speech involves the control of dynamic vocal tract postures using a high-dimensional neuromuscular system. Take the example of the human tongue. It comprises at least eight major muscle groups and behaves as a muscular hydrostat, with a constant volume structure that generates its own deformations under muscle contraction \citep{takano-honda2007, gick-etal2013, stone-etal2018}. A direct consequence of this organization is an enormous number of degrees of freedom. The tongue's muscle elements can be controlled with considerable independence; for example, the genioglossus is partitioned into quasi-independent neuromuscular compartments \citep{wrench2024}, which further multiplies the tongue's effective degrees of freedom. This poses a fundamental question:~how do speakers control a system with very high degrees of freedom to reliably generate the dynamic articulatory movements that underpin speech communication?

Despite its potential complexity, speech production appears to draw on a comparatively small repertoire of tongue postures \citep{harshman-etal1977, boe-etal1992, honda1996, perrier-etal1996, iskarous2005, story2005, calka-etal2024}. This connects to a widespread observation that motor control often operates at fewer effective degrees of freedom than are physically available \citep{bernstein1967}. That is, a smaller set of variables coordinate the many underlying muscles into functional configurations that can be combined to generate stable movement patterns. Speech appears to be no exception:~the small, consistent repertoire of tongue postures suggests that the vocal tract is also driven by fewer variables than its anatomy makes available \citep{gick2016}. As such, most models of articulatory control assume that the control of speech operates on some low-dimensional variables. Yet there remains a range of potential choices for this control basis, including tract variables, privileged articulatory dimensions, and statistical modes. These different variables often originate in different research traditions with distinct scientific goals, but they remain intimately related as different measurements of the same underlying system.

In this study, we aim to provide a new perspective on the control of tongue movement that identifies a potential bridge between these different research traditions. Specifically, we unify data-driven approaches to dimensionality reduction with formal dynamical models of articulatory control, where subsequently derived measurements represent a projection from this more holistic control space to a new coordinate system. In doing so, we hope to strengthen the connections between different approaches to the control of speech and identify a potential route toward more fully exposing the embodied nature of speech production.

\subsection{The dimensionality of vocal tract movements}
\label{subsec:dim}

Dimensionality reduction is a necessary feature of phonetic models, phonological theories, and quantitative methodology. It is often not possible or desirable to describe speech in terms of the full high-dimensional neuromuscular system, so a common approach is to segment the vocal tract into different anatomical structures, such as lower lip, tongue tip, tongue body, jaw, etc. Theories of speech production, such as Articulatory Phonology and Task Dynamics (AP/TD), commonly deploy this kind of dimensionality reduction \citep{fowler-etal1980, kelso-etal1986, saltzman-munhall1989, browman-goldstein1992}. For example, AP/TD distinguishes between \emph{task space} and \emph{articulator space}, where task space concerns one-dimensional vocal tract tasks, such as Lip Aperture (LA), Tongue Tip Constriction Degree (TTCD) or Tongue Body Constriction Location (TBCL). These tasks are associated with an articulator space, which comprises individual articulatory organs, such as tongue tip, tongue body, upper lip, lower lip, and so on. When an articulator such as the jaw is perturbed, the tongue can compensate to achieve intended vowel articulations \citep{gay-etal1981}, with a re-weighting of the relevant articulators to achieve a stable goal in task space. The same logic theoretically extends to the muscular level, where groups of muscles can be functionally combined to flexibly achieve articulator configurations.

Segmenting the vocal tract into distinct anatomical parts is a form of what \citet{gick2016} calls \emph{anatomical modularization}. Anatomical modularization is incredibly useful; it allows us to communicate concepts effectively and commands considerable descriptive adequacy, but it is not the only choice of modularization. For example, \citet{gick2016} contrasts this with \emph{functional modularization}, where the units of description are neuromuscular modules:~pre-compiled sets of co-activating muscles that combine to achieve functional behaviour \citep{bizzi-cheung2013, gick-stavness2013}. If speech is organized in this way, the resulting kinematics should show correspondingly low-dimensional structure. This motivates a substantial literature identifying empirical vocal tract \emph{modes}, which account for the major components of variance in complex articulatory data, but are kinematic rather than neuromuscular in character.

An influential statistical approach to vocal tract modes is the PARAFAC model of \citet{harshman-etal1977}, which decomposes tongue shapes for vowels into two speaker-independent factors with speaker-specific weights. These solutions can be language-specific \citep{jackson1988} but can also generalize across speakers and languages \citep{nix-etal1996}. Comparable low-dimensional solutions have been obtained from EMA \citep{hoole1999} and from 3D tongue surfaces \citep{zheng-etal2003}. Using PCA, two shape modes likewise account for most of the variance in 2D area functions derived from MRI \citep{story-titze1998, story2005}.

A related approach is model-based decomposition, which combines statistical modes with anatomical knowledge. This typically involves specifying an articulatory variable as a loading factor, such as jaw position, and then extracting the remaining components from the residuals \citep{maeda1990, maeda1991}. This approach has been extended to larynx and lip parameters \citep{beautemps-etal2001}, to three dimensions \citep{badin-etal2002}, and to multi-speaker MRI \citep{serrurier-etal2019}. The difference is consequential for what the resulting basis can express:~a purely statistical decomposition absorbs the jaw's contribution into its tongue modes, whereas a guided decomposition retains the jaw as a separable degree of freedom.

A compelling feature of shape modes is a potential link to underlying biomechanical mechanisms. Low-dimensional tongue shape patterns correspond closely to anatomical organisation \citep{honda1996} and have been related to acoustic output \citep{maeda-honda1994}, while principal components of the tongue surface recover deformations interpretable in muscular terms \citep{stone-etal1997, slud-etal2002}. Biomechanical simulations also show that the mechanics of a muscular hydrostat constrain the tongue to a small set of deformation patterns, even when its muscles are driven independently \citep{boe-etal1992, perrier-etal1996, payan-perrier1997}.

While functional accounts differ in how they identify the primary modes of tongue shape, they converge on the idea that observable behaviour reveals evidence of low-dimensional structure. This structure is not confined to static postures:~\citet{iskarous2005} shows that the dynamic control of tongue movement can be decomposed into qualitative patterns, while other studies have detailed the temporal dynamics of statistical modes \citep{story2007, ramanarayanan-etal2013}. Yet shape modes have seldom been treated as the coordinates in which a control law is formulated. This raises the question of which coordinates a model of articulatory control could actually use, which we address in the following section.

\subsection{A dynamical model for tongue shape dynamics}

A dynamical model of articulatory control contains three components:~(1) a \emph{change rule}, which is a mathematical equation that governs how vocal tract tasks unfold over time; (2) \emph{parameters}, which specify quantitative differences in movement within the constraints of the change rule; (3) a \emph{coordinate system}, which is the basis on which the control law operates. An example is the task dynamic model outlined by \citet{saltzman-munhall1989}, where (i) the change rule is a linear damped point attractor, in which vocal tract tasks evolve to an equilibrium position; (ii) the parameters are target, stiffness and damping values for each task; (iii) the coordinate system is tract variable space, with an additional nonlinear mapping to articulator space. This is an elegant model and it has subsequently been developed in order to better capture nonlinear movement characteristics, such as time-to-peak velocity and the relationship between temporal dynamics and movement amplitude. These extensions include reformulating the change rule to be nonlinear \cite{sorensen-gafos2016, kirkham2025, stern-shaw2025}, adding time-varying activation \citep{byrd-saltzman1998, tilsen2020}, and introducing time-varying parameters \cite{elie-etal2023}. Despite considerable theoretical development on these models, the choice of coordinate system has attracted much less attention. The majority of dynamical models use tract variables or articulators, but the studies reviewed in Section \ref{subsec:dim} suggest that alternative coordinate systems may be a viable prospect for low-dimensional control laws.

\begin{figure*}[t]
    \centering
    \includegraphics[width=\textwidth]{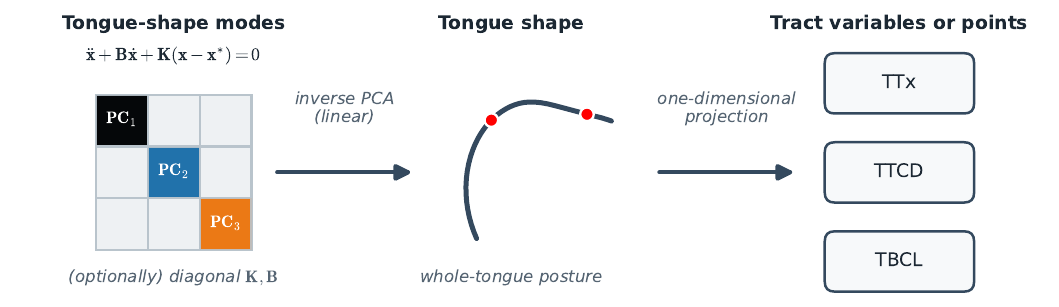}
    \caption{The proposed model architecture, where tongue shape modes are the control basis that drive tongue shape dynamics. One-dimensional measurements therefore represent a projection from tongue shape space into a new coordinate system.}
    \label{fig:coord_systems}
\end{figure*}

In this study, we build upon mathematical and theoretical advances across decades of research in task dynamics, but explore an alternative control basis in which tract and articulatory variables represent a projection from a set of more holistic control primitives. Our proposal is to extend the state space of current dynamical models from uncoupled tract variables to $N$-dimensional tongue (or vocal tract) shape features, which we denote $\mathbf{x} = \{x_1, x_2, \dots, x_n\}$. We assume that the dimensionality of $\mathbf{x}$ is low (i.e.~2--4 features) and that these features represent the principal dimensions of tongue shape. At this stage, we only assume that there exist a small number of dimensions that govern variation in the shape of the tongue and/or vocal tract. We illustrate this in Figure \ref{fig:coord_systems}.

Once the low-dimensional tongue shape features have been identified, their dynamics can be modelled as a dynamical equation. If we assume that the tongue shape features accurately represent variation in the shape of the tongue then the dynamical equation is fundamentally modelling whole tongue shape dynamics using low-dimensional control variables. We predict that the dynamics of $N$ tongue shape features for a given gesture can be governed by the damped linear point attractor,

\begin{equation}
\ddot{\mathbf{x}} = -\mathbf{K}(\mathbf{x} - \mathbf{x}^{*}) - \mathbf{B}\dot{\mathbf{x}}
\end{equation}

where $\mathbf{x}^{*}$ is an $N$-dimensional vector for the target or equilibrium position, $\mathbf{K}$ is a matrix of stiffness values, and $\mathbf{B}$ is a matrix of damping values. Each matrix contains a damping or stiffness value for each tongue shape mode, which can either be independent or can potentially interact or couple between modes. So far, these are the same quantities in previous task dynamics models \citep{saltzman-munhall1989}, but we also wish to consider an alternative to the common distinction between task and articulatory space. Specifically, we explore a model where tongue shape modes are the primitive control variables, which generate a holistic tongue shape. Tract or articulatory variables can then be viewed as a projection from tongue shape into a new coordinate system. Note that we are not proposing that we abandon the distinction between task and articulator space; our aim here is to identify a more holistic set of tongue shape modes that can be related to constriction-related tasks and multi-dimensionality articulator space.

\begin{figure}
    \centering
    \includegraphics[width=1\linewidth]{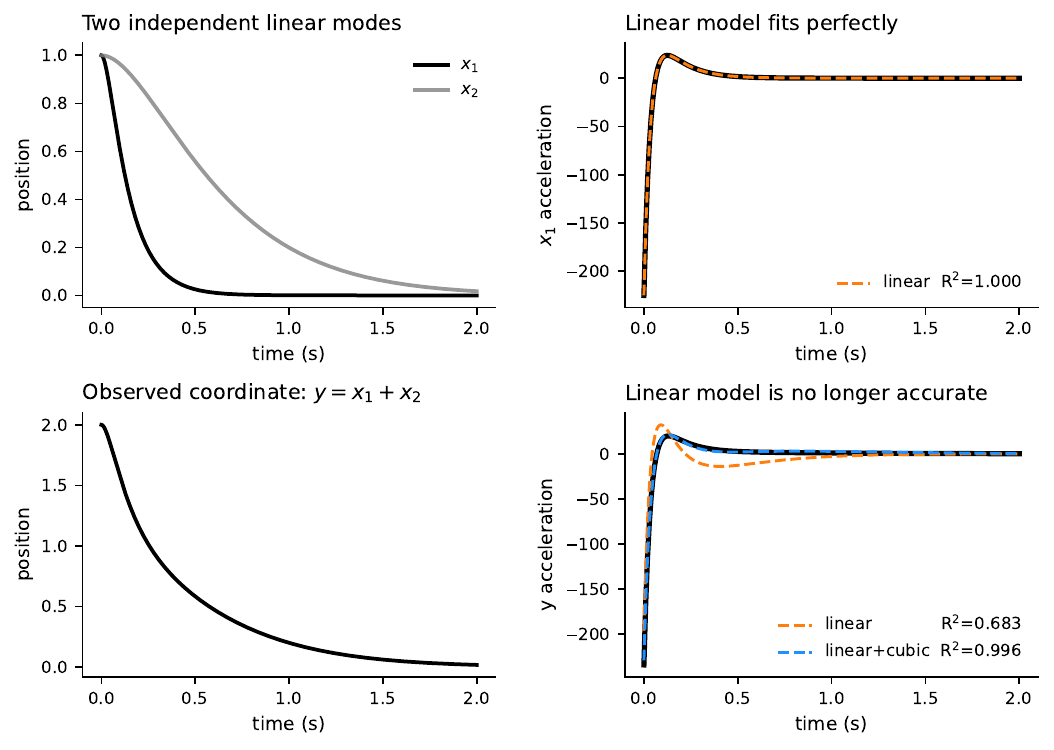}
    \caption{Mode mixing can manufacture apparent nonlinearity. Top left: two independent linear modes $x_1, x_2$. Top right: the acceleration of $x_1$ is captured essentially perfectly by a linear law ($x_2$ likewise, not shown). Bottom left: their linear superposition $y = x_1 + x_2$. Bottom right: the acceleration of $y$ is poorly fit by a linear model and can be fitted using a cubic term even though the underlying system is linear.}
    \label{fig:mode_mixing}
\end{figure}

One of the implications of our approach is that tongue shape modes generate a holistic dynamic tongue shape, from which we can then derive single point measurements for different parts of the tongue. But what happens when our tongue shape system is observed through a one-dimensional perspective, such as a tract variable? Figure \ref{fig:mode_mixing} shows a simple example. The top left panel shows two independent modes $x_1, x_2$ and the top right shows the acceleration trajectory for $x_1$, which is perfectly approximated by a linear model (note that this is also the case for $x_2$, not shown here). The bottom left panel shows the linear combination of the two modes $y = x_1 + x_2$ and the corresponding acceleration trajectory in the bottom right, where the linear model is no longer a good fit ($R^2 = 0.683$) due to the different timescales of $x_1$ and $x_2$. The apparent nonlinearity is thus a property of the fitted model rather than the underlying system. This is an instance of a general result:~projecting a system onto fewer variables does not preserve the form of the governing law \citep{zwanzig1961, mori1965}. Fitting a second-order law to such a projection therefore imposes only a partial model of the underlying dynamics, where residual structure can be mistaken for nonlinearity. This is what we propose may be happening when we analyse one-dimensional variables, such as tract variables or simple articulatory coordinates. Specifically, these measurements are projections from the higher-dimensional coordinate space of tongue shape, with the result that their dynamics may show behaviour that is not a true property of the underlying control system. Note that the same is potentially true of gestural blending in Articulatory Phonology, which represents a different form of mode mixing, so this observation is a more general consideration for any model of articulatory control.

We exercise caution regarding the limits of the above argumentation, because a mixture of linear modes is itself a linear system and scaling the amplitude of a movement scales the whole trajectory while leaving its time course unchanged. As a result, this mode mixing cannot, on its own, generate amplitude-dependent kinematics of the kind that motivate nonlinear control laws, such as the nonlinear relation between movement amplitude and time to peak velocity \citep{ostry-munhall1985, sorensen-gafos2016}. Such phenomena would require genuine nonlinearity, such as nonlinear coupling between modes. Our claim is narrower:~an apparent departure from second-order linear dynamics in a one-dimensional variable does not by itself motivate a nonlinear control law, because such departures arise generically from projection between different coordinate systems.

\subsection{This study}

In the sections that follow, we implement a quantitative approach to fitting and interpreting a linear dynamical of tongue shape modes. We fit this model to ultrasound tongue imaging data on monophthongal vowels from British English speakers. This allows us to test the approach using a relatively simple set of controlled movements, with a focus on the vowel formation and its release into a coronal consonant. If successful, this would motivate extending the model to more complex tongue movements, such as diphthongs and consonant constrictions. In summary, our aim is to evaluate whether a linear dynamical model of tongue shape variables is sufficient for capturing lingual shape dynamics in vowels.

\section{Methods}

\subsection{Data}

We used the TarDiS corpus \citep{strycharczuk-etal2025,strycharczuk-kirkham2025}, which comprises 40 speakers from the North West of England \citep{strycharczuk-etal2020b} recorded using midsagittal ultrasound. We focus on a subset of the corpus; namely short monophthongs \emph{bid, bed, bad, bud, bod}, embedded in the carrier phrases ``She says X'' and ``She says X eagerly''. All vowels were repeated 4--6 times per speaker. We focus on short vowels in this study in order to narrow the focus to a tractable set of simple vowel gestures. Our previous work shows that long vowels and diphthongs typically require two gestural activation intervals in English \citep{strycharczuk-etal2024}, which would require a more complex piecewise model and accurate identification of target boundaries from multi-dimensional space. We discuss prospects for such an analysis in \S\ref{subsec:limitations}.

The acoustic signal was labelled for C1 (beginning at the onset of bilabial closure), V (onset of modal voicing), and C2 (from onset of stop closure until stop release). We use the entire [C$_{1}$VC$_{2}$] interval for analysis; specifically, C1 to V midpoint represents the formation of the vowel target gesture, whereas V midpoint to C2 represents the release of the vowel target into the coronal gesture for /d/. We discuss gesture segmentation in Section \ref{subsec:segmentation}. Four speakers were excluded due to contour-tracking artefacts (MCR6, MCR8), anomalously low PC\textsubscript{1} amplitude with a few outlier tokens (MCR17) and release-phase segmentation issues (MCR27). This resulted in 36 speakers for analysis in the final analysis, with a total of 1687 tokens.

\subsection{Tongue shape representation}
\label{subsec:data_processing}

Tongue shape was parameterized by estimating anatomically-defined points representing midsagittal tongue shape, which was computed using a pre-trained model \citep{wrench-balch-tomes2022} implemented in DeepLabCut (DLC) \citep{mathis-etal2018}. The DLC points were then subjected to Generalized Procrustes Analysis (GPA), which removes translation, rotation and scaling differences between speakers, followed by Principal Components Analysis (PCA) in the tangent space \cite{kent1992}. Our GPA procedure scales all speakers to a common size and rotation, but does not fundamentally alter tongue shape; see \citet{lo-etal2025, strycharczuk-etal2026} for further information and validation of this approach.

The PCA on GPA-aligned contours yields an empirically discovered low-dimensional representation of whole-tongue shape, which we treat as hypothetical control variables in a dynamical model. We retain the first three principal components as the state variables for the dynamical model, which together represent 75.1\% of cumulative variance. GPA-PCA necessarily lowers the per-component variance explained compared with by-speaker PCA, because it must also span between-speaker baseline shape differences, which largely exist in the higher PCs. In order to verify that our GPA pooling is representative of individual speaker behaviour, we also fitted by-speaker PCA (mean-centered) and calculated the variance captured by the first 3 PCs for each speaker. We compared the by-speaker PCA variance to the GPA-PCA variance for each speaker and find that the shared GPA modes retain 93\% of variance from the by-speaker PCA (range 0.59--0.96). This confirms that a shared coordinate system largely captures the same variability as a by-speaker PCA, with the advantage of a shared cross-speaker coordinate space for tongue shape variation.

\begin{figure*}
    \centering
    \includegraphics[width=1\textwidth]{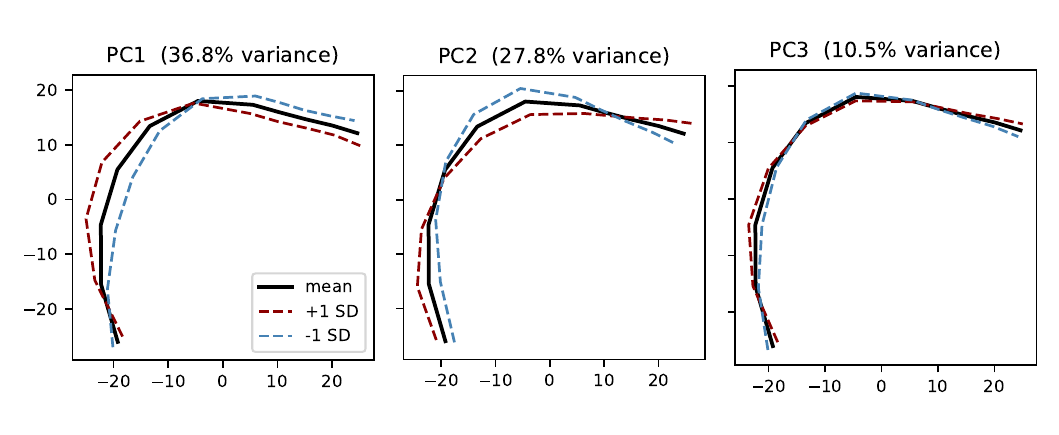}
     \includegraphics[width=1\textwidth]{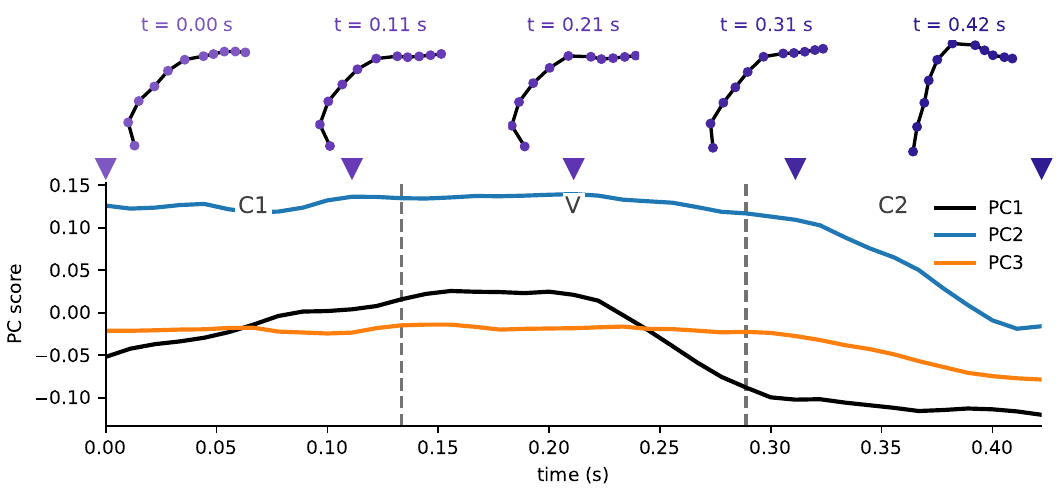}
    \caption{Left: Principal component loadings, with the mean midsagittal tongue shape and $\pm 1$\,SD perturbation. Right: Time-varying principal components in \emph{bad}, with midsagittal tongue contours at evenly spaced times across the [C$_{1}$VC$_{2}$] window.}
    \label{fig:pc_interp}
\end{figure*}

The interpretation of the three PCs is visualized in Figure~\ref{fig:pc_interp} (top). PC\textsubscript{1} corresponds broadly to front-back tongue body position, PC\textsubscript{2} to high-low body position, PC\textsubscript{3}  to a finer tip/blade contrast. We note that fitting a dynamical model to two versus three PCs result in very similar models, with almost identical performance, suggesting that PC\textsubscript{3} is not particularly useful in distinguishing vowel categories. Despite this, we retain three PCs in our subsequent modelling because PC\textsubscript{3} still accounts for 10\% of variance and it represents a more complete representation with some information on tongue tip movement, which may be useful in extending the model to a wider range of movements.

We obtain PCs for each ultrasound frame across the [C$_{1}$VC$_{2}$] window.  Figure~\ref{fig:pc_interp} (bottom) shows an example of time-varying PCs for one token of \emph{bad}, with raw tongue contours sampled at equidistant points across the time window. Per-token PC position trajectories were smoothed using a third-order polynomial Savitzky-Golay filter across 7-sample windows. We discuss the sensitivity of the estimated parameters to smoothing windows in Section \ref{subsec:freq-damp}.

\subsection{Dynamical model fitting}

The dynamics are modelled as a linear damped point attractor; namely, a piecewise damped harmonic oscillator. This is the same model form that is used in linear task dynamic models of tract variables \citep{saltzman-munhall1989}. As a closed-form solution of this equation exists, we fit the analytic solution directly rather than numerically integrating the dynamics, which permits a simple least-squares fit,

\begin{equation}
\ddot{\mathbf{x}} = -\boldsymbol{\Omega}^2(\mathbf{x}-\mathbf{x}^*)
- 2\mathbf{Z}\boldsymbol{\Omega}\,\dot{\mathbf{x}},
\label{eq:msd}
\end{equation}

where $\boldsymbol{\Omega} = \operatorname{diag}(\omega_i)$ and $\mathbf{Z} = \operatorname{diag}(\zeta_i)$. The treatment of $\boldsymbol{\Omega}$ and $\mathbf{Z}$ as diagonal means we model each component independently, with the system therefore reducing to three independent oscillators of the form

\begin{equation}
\ddot{x}_i = -\omega_i^2 \left( x_i - x^*_i \right)
- 2 \zeta_i \omega_i \dot{x}_i ,
\label{eq:dho_component}
\end{equation}

where $i = 1,2,3$ represent the first three principal components. The fitted parameters are the target $x^*_i$, damping ratio $\zeta_i$ and natural frequency $\omega_i$ for each component, with initial position and velocity supplied from empirical estimates. Note that $\omega$ and $\zeta$ map onto the conventional stiffness and damping coefficients as $k = m \omega^2$ and $b = 2 m \zeta \omega$, given an effective mass $m$.

\subsection{Fitting models to windowed segments}
\label{subsec:segmentation}

Each [C$_{1}$VC$_{2}$] window was partitioned at the vowel midpoint into a \emph{closure} phase (C1 release into V target) and a \emph{release} phase (V release into C2 target). A separate oscillator was fitted to each phase independently. We sensitivity tested the segmentation point between closure and release, because it was not possible to derive a stable landmark based on a 3-dimensional noisy velocity signal from the principal component time-series. Specifically, we tested closure/release split points at 40\%, 50\%, and 60\% of the acoustic vowel duration. Median closure/release RMSE (root mean square error) was 0.163/0.312~mm (40\%), 0.177/0.286~mm (50\%), and 0.209/0.247~mm (60\%). These values are uniformly low, with RMSE decreasing for closure and increasing for release as the segmentation point is increased. In practice, the fits are relatively insensitive to the precise segmentation point within this bound, so we chose to segment at the vowel midpoint (50\%), which is a reasonable proxy for the timing of the vowel target in short monophthongs. Figure~\ref{fig:pc_interp} (right) shows a representative example, where the V label represents the segmentation point between closure and release for this token.

\subsection{Parameter estimation and evaluation}
\label{subsec:param-est}

\begin{figure*}
    \centering
    \includegraphics[width=1\textwidth]{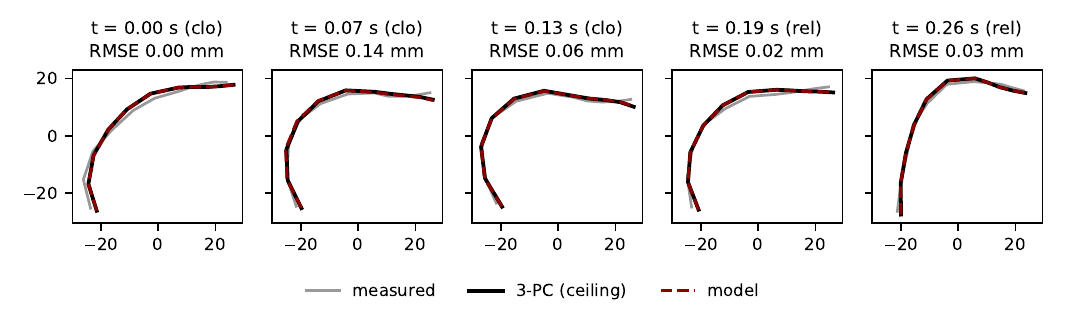}
    \caption{Tongue contour reconstruction for \emph{bod}. Each panel is a selected frame across the [C-\textsubscript{1}V-C\textsubscript{2}] window, with raw contour (grey), 3-PC reduction (black), model prediction (red). RMSE is the difference between model prediction and 3-PC input.}
    \label{fig:reconstruct_best}
\end{figure*}

We constrained parameter estimation, with $\zeta$ estimated from a 60-point search over [0.05, 3.0], where 1.0 is critically damped, and $\omega T \in [0.05, 500]$ where $T$ is the duration of the fitted window. Model fit was evaluated as the reconstruction error of each oscillator fit against its 3-PC input, expressed in the original tongue contour measurement space (in millimetres). Specifically, the RMSE PC-score residual is rescaled to physical units as

\begin{equation}
\text{RMSE}_\text{mm} = \text{RMSE}_\text{pc}\times\tilde{S}\sqrt{\frac{n_\text{pc}}{n_\text{L}}},
\label{eq:rmse}
\end{equation}

where $\tilde S$ is the median GPA centroid size, $n_\text{pc}$ is the number of principal components, and $n_L$ is the number of tongue-contour landmarks. Our reconstruction error only concerns the 3-PC reduction of tongue contour shape, as the model cannot be expected to predict higher-dimensional features that are not part of the input. Figure~\ref{fig:reconstruct_best} illustrates this for a token of \emph{bod}, where each subplot represents a sample frame across [C$_{1}$VC$_{2}$], with three contours:~the raw tongue contour (grey), the 3-PC reduction (black), and the oscillator forward simulation (red dashed). The 3-PC reduction is the target contour that we aim to model, so this is the theoretical ceiling for model accuracy. The raw contour in grey demonstrates the extent of PCA variance loss. In this case, the model-vs-3-PC input has RMSE of only 0.00--0.14\,mm, while the total RMSE (model vs. measured) ranges 0.73--1.52\,mm. The model therefore reaches the three-PC ceiling almost exactly, and the residual disagreement with the raw contour is a consequence of the PCA reduction rather than model fit error.

An important caveat to emphasize is that we score model fit only on position, because the velocity estimates from PC trajectories were noisy and require a degree of smoothing that could overly bias the observed structure. This has consequence for interpretation. Position trajectories are relatively insensitive to the order of the dynamics that generate them; for example, see \citet{kirkham2025b} for an example of first-order dynamics that look appropriate in position space, but are inadequate in velocity and acceleration. Reconstruction accuracy in position is therefore evidence that the model is \emph{sufficient} for these trajectories, rather than conclusive evidence that the underlying control law is linear or second-order. We interpret our model fits accordingly.

We also report additional statistics beyond model fit in order to provide a measure of parameter interpretability. We focus on the target value $x^*$, which we predict should show a strong association with vowel category. As we fit to closure (vowel target) and release (coronal target), we also predict that coronal targets should cluster more closely and show less vowel-specific effects, whereas vowel targets should be more distinct. We quantify this using the multivariate Fisher ratio $\sqrt{\operatorname{tr}(W^{-1}B)}$ where $W$ and $B$ are the within- and between-category scatter matrices of the equilibrium targets in \{PC\textsubscript{1}, PC\textsubscript{2}\} space. This metric is typically maximized in methods such as linear discriminant analysis and is a multivariate measure of class separability. Values increasingly above 1 indicate increased separability. Finally, $V > D$ is a metric of how many speakers have higher $\text{sep}_V$ than $\text{sep}_D$, which is calculated on by-speaker medians for closure and release.

\section{Validation experiment}
\label{subsec:validation}

\subsection{Method}

Before fitting models to empirical data, we first validated whether dynamical parameters can be recovered from synthetic data. This allows us to separate the quality of model fit from the identifiability of its parameters. Specifically, it is possible that there could be a large set of parameter values that could fit a single trajectory; if so, then any estimates from empirical data should be viewed as one plausible set of estimates, rather than an estimate of underlying ground truth. We simulated three components spanning underdamped, critically damped and overdamped regimes $\zeta \in [0.5, 1.0, 1.5]$ at different natural frequencies $\omega \in [30, 45, 55]$~rad/s, with fixed targets, initial positions and initial velocities. Ground-truth trajectories were obtained by numerically integrating the oscillator and sampling at the frame rate in our empirical ultrasound data (82~Hz). We simulated window durations over 80--300 ms, spanning the empirical range (median 207~ms, IQR 171--256~ms), and added Gaussian noise to position with a standard deviation of 2.5\% of each component's range. All trajectories passed through the same smoothing pipeline in Section \ref{subsec:data_processing}, and we then fitted the analytical model. An important aspect of the model is that initial velocity is an input, which must be estimated for empirical data. We thus compared two conditions:~(1) true initial velocity from the synthetic data (\emph{exact}); (2) an estimate of initial velocity from the smoothed positions as in the empirical pipeline (\emph{estimated}). We quantify parameter recovery as the signed error in $\omega$, $\zeta$ and $x^*$ relative to ground truth.

\subsection{Results and summary}

\begin{figure}
    \centering
    \includegraphics[width=1\linewidth]{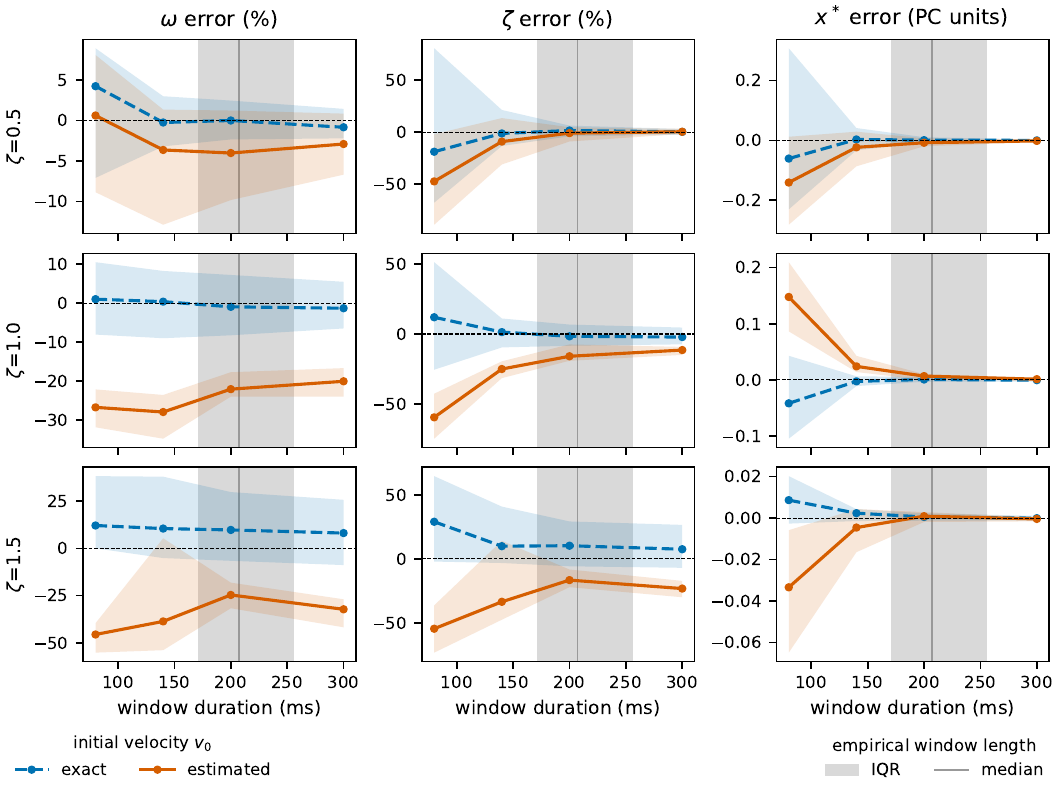}
    \caption{Parameter-recovery error as a function of window duration across damping ratio (rows) and parameter (columns).}
    \label{fig:validation}
\end{figure}

Figure \ref{fig:validation} shows parameter recovery accuracy across conditions, where the grey banded area represents the interquartile range of empirical durations and the vertical line is the median duration. We find that $\omega$ in the exact velocity condition has very low error except for a median offset of +9.8\% when $\zeta = 1.5$, while the estimated velocity condition systematically underestimates $\omega$. Damping is generally accurate for $\zeta = 0.5$, but less accurate for higher damping ratios, where the estimated velocity conditions has $\sim$16\% median error. The target $x^*$ is close to ground truth within the empirical duration window in both conditions. These results show that estimated initial velocity degrades parameter recovery, with up to $\sim$25\% error for $\omega$ and up to $\sim$16\% inaccuracy for $\zeta$. This is likely an under-estimate compared with real data, which is noisier and more variable than our synthetic data, leading us to conclude that damping ratio and frequency parameters are likely to have moderate-to-significant levels of inaccuracy. In our analysis, we address this in two ways:~(1) we explicitly test the effect of constraining model parameters to improve interpretability and identifiability; (2) we treat any estimates of damping and frequency as unreliable estimates from an under-determined system.

\section{Empirical experiment}

\subsection{Overview}
\label{subsec:fit}

\subsection{Model fit and reconstruction accuracy}
\label{subsec:fit}

We fitted a piecewise linear damped point attractor to time-varying principal components representing tongue shape, with separate models for vowel target and release phases. The model comparison is shown in Table \ref{tab:model_comparison}. For each model we calculated by-speaker medians and then report the sample-level median across the 36 speakers, with the interquartile range for between-speaker variation.

\begin{table*}[htbp]
  \centering
  \footnotesize
  \caption{Model comparison for a damped harmonic oscillator fitted to $\mathrm{PC}_{1-3}$ tongue contours. Unconstrained represents a model with free parameters for $x^*, \zeta, \omega$. Each cell is the sample median with the interquartile range in brackets; RMSE in millimetres, clo/rel = closure/release. $V > D$ counts speakers with $\text{sep}_V > \text{sep}_D$.}
  \label{tab:model_comparison}
  \begin{tabular}{l @{\hspace{1em}} l @{\hspace{1em}} l @{\hspace{1em}} l @{\hspace{1em}} l @{\hspace{1em}} l}
    \hline
    Model & $\text{RMSE}_{\text{clo}}$ & $\text{RMSE}_{\text{rel}}$ & $\text{sep}_V$ & $\text{sep}_D$ & $V > D$ \\
    \hline
    unconstrained & 0.177 [0.157, 0.221] & 0.286 [0.248, 0.327] & 0.57 [0.47, 0.88] & 0.51 [0.46, 0.68] & 23/36 \\
    + $\mathrm{PC}_{1}$ $\zeta_{\mathrm{critical}}$ & 0.220 [0.182, 0.263] & 0.384 [0.303, 0.452] & 0.52 [0.48, 0.63] & 0.52 [0.43, 0.63] & 19/36 \\
    + $\mathrm{PC}_{1-3}$ $\zeta_{\mathrm{critical}}$ & 0.263 [0.218, 0.322] & 0.472 [0.372, 0.592] & 0.50 [0.45, 0.65] & 0.52 [0.44, 0.60] & 18/36 \\
    + pin $x^*$ & 0.230 [0.191, 0.296] & 0.407 [0.338, 0.505] & 5.14 [4.00, 6.29] & 0.52 [0.31, 0.73] & 36/36 \\
    + pin $x^*$ + $\mathrm{PC}_{1}$ $\zeta_{\mathrm{critical}}$ & 0.303 [0.250, 0.379] & 0.531 [0.433, 0.707] & 5.14 [4.00, 6.29] & 0.52 [0.31, 0.73] & 36/36 \\
    + pin $x^*$ + $\mathrm{PC}_{1-3}$ $\zeta_{\mathrm{critical}}$ & 0.357 [0.325, 0.462] & 0.647 [0.539, 0.869] & 5.14 [4.00, 6.29] & 0.52 [0.31, 0.73] & 36/36 \\
    \hline
  \end{tabular}
\end{table*}

The unconstrained model achieves very high reconstruction accuracy, with RMSE of 0.177 mm (closure) and 0.286 mm (release). However, the model has very poor separability of vowel targets, with median $\text{sep}_V$ = 0.57, which is only marginally higher than $\text{sep}_D$ = 0.51, both of which are substantially below a separability threshold of 1. Only 23/36 speakers have $\text{sep}_V > \text{sep}_D$ meaning that vowel targets are not well recovered. Further inspection shows that the fitted targets are treated as free parameters and do not have a straightforward interpretation.

We subsequently tested whether model interpretability could be improved by constraining the damping ratio to critical $\zeta = 1$ for $\mathrm{PC_{1}}$-only and also for $\mathrm{PC_{1-3}}$. This resulted in a small increase in RMSE for $\mathrm{PC_{1}}$ of +0.043 mm (closure) and +0.098 mm (release), and an increase for $\mathrm{PC_{1-3}}$ of +0.086 (closure) and +0.186 (release). Yet $\text{sep}_V$ slightly decreases in both conditions, which makes it a worse model on all accounts. Specifically, critical damping appears to degrade phonetic target recovery further, evidenced in the decrease in the number of speakers for whom $\text{sep}_V > \text{sep}_D$ (from 23/36 to 19/36 for $\mathrm{PC_{1}}$ and 18/36 for $\mathrm{PC_{1-3}}$).

As constraining damping to critical is insufficient for recovering interpretable target parameters, we test an additional constraint, where the target $x^*$ is pinned to the segment endpoint value. This means that the target is no longer estimated, but is treated as a fixed parameter separately for each closure and release phase for each token. This only results in a small increase in RMSE compared with the unconstrained model, with +0.053 mm (closure) and +0.121 mm (release), but a substantial increase in $\text{sep}_V$ = 5.14 alongside a low $\text{sep}_D$ = 0.52. All conditions with pinned $x^*$ show $\text{sep}_V > \text{sep}_D$ for all speakers, because $\text{sep}_V$ and $\text{sep}_D$ do not vary between the pinned conditions. This is worth emphasising, as we are no longer fitting the target value in the model, but pinning it at the endpoint of a segmentation window. Nonetheless, this reduces the number of parameters that the model has to fit and has only a small detrimental effect on fitting accuracy.

If we further constrain the pinned-target model with the addition of critical damping then RMSE increases by +0.073 mm (closure) and +0.124 mm (release) when $\mathrm{PC_{1}}$ is constrained, which increases to +0.127 mm (closure) and +0.240 mm (release) when all PCs are constrained to critical damping. However, by the time $x^*$ is pinned and all PCs are critically damped, RMSE has increased by 102\% (closure) and 126\% (release), meaning that fitting error has more than doubled. This reveals a clear trade-off between model fit and parameter interpretability; e.g.~while the most constrained model has the highest RMSE, it only has a single fitted $\omega$ parameter per-PC, alongside a fixed target value. As our aim is not to discover a single `true' model, we do not arrive at a final best model, but in subsequent sections we focus on the pinned-target models as these are clearly more interpretable than the unconstrained models.

\subsection{Target parameter}
\label{subsec:vquad}

\begin{figure*}
    \centering
    \includegraphics[width=1\textwidth]{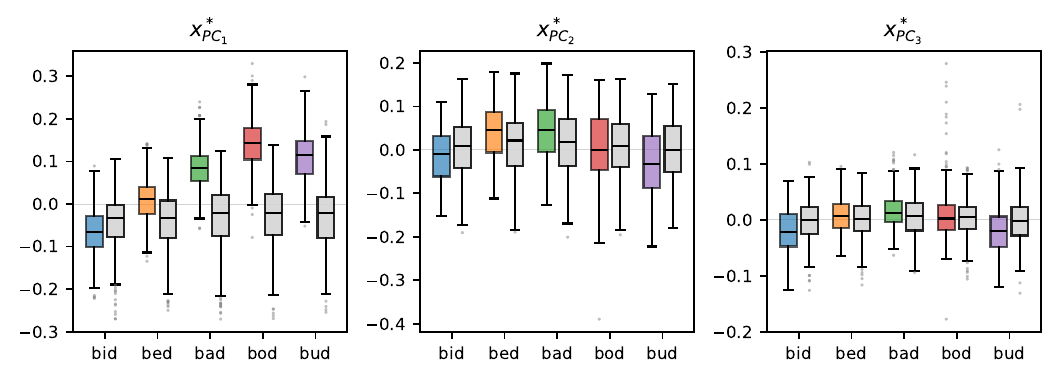}
    \caption{Distributions of the target $\mathbf{x}^*$ for PC\textsubscript{1}--PC3 across vowel closure (colours) and release (grey).}
    \label{fig:xeq_boxplots}
\end{figure*}

We now turn to interpreting the dynamical parameters of the model. We begin by focusing on the target parameter $x^*$, which is specified independently for each tongue shape feature. How well do these discovered targets correspond to phonological vowel categories? Recall that the unconstrained model fits did not yield interpretable target parameters, so we pinned to  $x^*$ to window endpoints, making targets a product of the data and segmentation, rather than model optimization. As such, the targets analysis better reflects the validity of tongue shape modes as a coordinate system, rather than the dynamical model per se. Figure \ref{fig:xeq_boxplots} shows the distribution of values for the target $x^*$ in \{PC\textsubscript{1}, PC\textsubscript{2}, PC\textsubscript{3}\}. We see excellent separation of vowel categories in $x^*_{\text{PC}_{1}}$, which is closely related to front/back differences. The grey boxes show the associated vowel release targets into /d/, which strongly cluster together. The results for $x^*_{\text{PC}_{2}}$ show smaller vowel-specific differences, reflecting subtle differences in vowel anteriority. $x^*_{\text{PC}_{3}}$ does not appear to systematically vary by vowel. See Figure \ref{fig:pc_interp} for interpretation of the PC modes.

In Section \ref{subsec:param-est}, we reported high separability of vowel targets in \{PC\textsubscript{1}, PC\textsubscript{2}\} space, based on the multivariate Fisher ratio. We subsequently calibrated this metric against a within-speaker label-permutation null hypothesis. Specifically, for each speaker and phase we computed the observed target separability in the \{PC\textsubscript{1}, PC\textsubscript{2}\} plane individually for closure and release phases, then built a null distribution by randomly re-assigning vowel labels to that speaker's own tokens 1000 times and recomputing separability. If a speaker's observed separability exceeded the 95th percentile of their null distribution this was considered to refute the null hypothesis of no separability. For the pinned-target model, all 36 speakers showed significant vowel separability in the closure phase (median $\text{sep}_V = 5.14$), against 31\% of speakers in the release phase (median $\text{sep}_D = 0.52$). This confirms that the kinematic vowel targets do carry significant vowel identity, whereas there is only limited evidence for vowel identity in the release phase, which likely represents coarticulation. For comparison, we conducted the same test for the unconstrained model, where only 56\% of speakers exceed the null for closure and 22\% for release.

\begin{figure}
    \centering
    \includegraphics[width=1\linewidth]{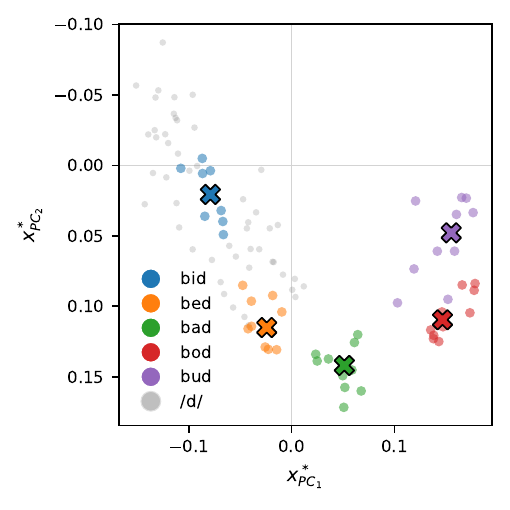}
    \caption{Target values in \{PC\textsubscript{1}, PC\textsubscript{2}\} space for speaker MCR14. Crosses represent by-vowel medians and points represent individual tokens. Axes have been oriented to align with conventions for representing vowels in $\{F\textsubscript{1} F\textsubscript{2}\}$ space.}
    \label{fig:xeq_mcr14}
\end{figure}

Figure~\ref{fig:xeq_mcr14} plots \{PC\textsubscript{1}, PC\textsubscript{2}\} across all tokens for one speaker (MCR14; $\text{sep}_V = 6.29$, $\text{sep}_D = 0.21$). The vowels cluster into a near-canonical vowel quadrilateral, with almost no overlap between tokens. The release targets cluster into the top left of the vowel space, suggesting that the target for /d/ is similar to a front vowel and in many cases higher and more anterior.

\subsection{Frequency and damping ratio}
\label{subsec:freq-damp}

In Section \ref{subsec:validation}, we showed that damping ratio and frequency cannot always be recovered with high accuracy in synthetic data, and should be treated as a property of the model fits rather than an estimate of underlying truth. Table \ref{tab:omega_zeta} shows $\omega$ and $\zeta$ values for the unconstrained and pinned $x^*$ conditions. The $\omega$ range of 22.5--28.5 rad/s corresponds to a frequency of 3.58--4.54 Hz, with both unconstrained and pinned models showing an increase in frequency from $\mathrm{PC_{1}}$ to $\mathrm{PC_{3}}$ for the closure phase. Release phase frequencies show a less clear pattern, but $\mathrm{PC_{1}}$ is faster than higher PCs in both cases. In terms of damping ratio, $\zeta$ is 0.55--0.66 (unconstrained) and 0.68--0.81 (pinned $x^*$) in the closure phase, which represents near-critically-damped dynamics when the target is pinned. In contrast, the release is always heavily under-damped with 0.16--0.38 (unconstrained) and 0.27--0.57 (pinned $x^*$). This may be a consequence of the acoustic segmentation truncating the true gestural target for /d/, with the fit compensating via a more extreme virtual target.

\begin{table*}[htbp]
  \centering
  \footnotesize
  \caption{Estimated $\omega$ (rad/s) and $\zeta$ for the unconstrained and pinned-target (+ pin $x^*$) models, as well as estimates for $\omega$ when damping is fixed to critical ($\omega_{\zeta_{\text{critical}}}$) for all PCs. Each cell is the median with interquartile range in brackets.}
  \label{tab:omega_zeta}
  \begin{tabular}{l @{\hspace{1em}} l @{\hspace{1em}} ll @{\hspace{1em}} ll @{\hspace{1em}} ll}
    \hline
     & & \multicolumn{2}{c}{$\omega$ (rad/s)} & \multicolumn{2}{c}{$\zeta$} & \multicolumn{2}{c}{$\omega_{\zeta_{\text{critical}}}$ (rad/s)} \\
    \cline{3-4}\cline{5-6}\cline{7-8}
    Model & PC & closure & release & closure & release & closure & release \\
    \hline
    unconstrained & 1 & 22.5 [18.3, 24.7] & 25.1 [21.9, 28.6] & 0.64 [0.50, 0.77] & 0.38 [0.28, 0.53] & 23.2 [17.1, 25.6] & 32.1 [26.1, 39.0] \\
                  & 2 & 24.1 [20.1, 29.4] & 20.5 [17.2, 24.8] & 0.55 [0.37, 0.90] & 0.16 [0.11, 0.26] & 25.6 [19.1, 30.4] & 12.8 [8.6, 35.3] \\
                  & 3 & 28.5 [25.0, 34.0] & 23.5 [20.7, 27.3] & 0.66 [0.55, 0.98] & 0.24 [0.17, 0.31] & 32.5 [24.9, 38.7] & 23.2 [16.5, 37.0] \\
    \hline
    + pin $x^*$  & 1 & 23.0 [18.8, 24.0] & 22.7 [19.5, 26.1] & 0.70 [0.65, 0.78] & 0.57 [0.39, 0.69] & 25.0 [20.3, 28.1] & 29.2 [22.4, 37.0] \\
                  & 2 & 22.7 [19.4, 26.6] & 15.4 [12.6, 16.9] & 0.68 [0.58, 0.78] & 0.27 [0.19, 0.37] & 25.6 [21.7, 30.3] & 14.0 [11.7, 18.8] \\
                  & 3 & 27.3 [23.1, 31.0] & 17.4 [14.6, 20.7] & 0.81 [0.68, 1.01] & 0.41 [0.31, 0.50] & 29.0 [23.9, 35.0] & 15.9 [12.4, 21.4] \\
    \hline
  \end{tabular}
\end{table*}

We checked whether the $\zeta_\text{critical}$ changes the recovered frequencies when all PCs are constrained to critical damping. Closure $\omega$ is only slightly higher under critical damping, which is unsurprising as damping was moderate-to-near-critical in unconstrained cases. The release dynamics are much more variable, which suggests that near-critical damping is likely not the appropriate parameterization of the release. As noted above, this could be an artefact of segmentation decisions, so we do not attach too much weight to this observation. Finally, we note that damping is heavily sensitive to smoothing. We sensitivity tested the Savitzky-Golay smoothing window across \{5, 7, 9\} samples and fit quality improved with window length:~$\text{RMSE}_{clo}$ = \{0.255, 0.177, 0.136\} mm. Damping ratio for the closure $\zeta$ ranged from 1.10 (5-samples) to 0.49 (9-samples), whereas the release phase remained heavily underdamped throughout, with $\zeta$ = \{0.26, 0.24, 0.21\}.  Natural frequencies varied less than damping; e.g.~median closure $\omega$ = \{30.3, 24.1, 23.3\} rad/s; release $\omega$ = \{25.1, 23.5, 22.8\} rad/s. While the qualitative pattern is fairly robust to smoothing, this strengthens our point that the frequency and damping ratio are better viewed as a property of the model fit, rather than estimates of true physical quantities.

\subsection{Summary of results}

The results show that a linear damped point attractor provides an excellent fit to a low-dimensional representation of tongue shape dynamics. This specifically means that the temporal evolution from a starting position to a target position can be handled entirely by the intrinsic, autonomous dynamics of a damped point attractor system. While the model is an excellent fit, the unconstrained model does not yield phonetically-interpretable targets. We found that interpretable targets were only achievable in the present modelling by explicitly specifying the target positions, which were estimated from the endpoints of the segmented windows. In terms of damping and frequency, the estimated parameters show the vowel formation gesture is near-critically damped, while the release gesture is heavily under-damped.

\section{Discussion}

\subsection{Shape modes as a potential basis for lingual control}

Our primary contribution is a potential view of articulatory control where holistic shape modes represent the control variables. We show that this has numerous theoretical advantages, including a cross-speaker coordinate space and an explanation for why projection of tongue shape into one-dimensional articulatory measurement can yield nonlinear artefacts. Our model is also entirely linear in control space, with modal variables that linearly combine to generate observed tongue shape dynamics. That said, we caution against over-interpreting these results regarding linearity, because we have only evaluated the position fits for tongue shape, whereas higher-order derivatives provide stronger signatures of (non)linearity \citep{kirkham2025b, sorensen-gafos2016}. In the present data, we opted against scoring the velocity and acceleration predictions, because the time-varying principal components are significantly noisier than EMA or flesh-point trajectories. In future work, it may be productive to analyse velocity and acceleration predictions from the model, yet in the absence of an empirical signal for comparison this will necessarily remain a qualitative evaluation. Another avenue would be more constrained prediction of the tongue contour dynamics from the ultrasound data to achieve better velocity estimates \citep{gubian-etal2015}.

\subsection{From tongue shape modes to muscle synergies?}

Our analysis shows that tongue shape modes are a viable control basis for modelling articulatory control, with tract or articulatory variables representing a projection from tongue shape space to a new coordinate space. Yet we predict that the use of functional modes may have deeper theoretical utility that could significantly advance our understanding of speech. A large literature in motor control has posited the concept of \emph{muscle synergies}, which represent co-activating muscle modules that linearly combine to generate functional patterns of behaviour \citep{mussa-ivaldi-etal1994, davella-etal2003}. We make the speculative prediction that empirical tongue shape modes may have a strong relationship with underlying muscle synergies. This presents a way of addressing a fundamental issue in task dynamics, which has -- for understandable reasons -- been largely neglected; that is, the mapping between high-dimensional muscle space and the lower dimensional space of control and observation. We note that previous work has attempted to similarly establish relationships between statistical control variables and muscle activations \citep{honda1996}, as well as acoustics \citep{maeda-honda1994}. Two lines of work are especially encouraging here:~\citet{sorensen-etal2019} recover task-dependent articulator synergies from a statistical kinematic model of MRI data, while \citet{woo-etal2021} apply sparse non-negative matrix factorization to tagged MRI to identify functional units of the tongue that deform coherently.

Given the challenges of recording muscle-level EMG data, the current generation of high-fidelity biomechanical models will be an essential tool in further testing the relationship between empirical modes and muscle synergies \citep{lloyd-etal2012, calka-etal2025}. Yet the dimensionality of synergies derived from muscle models are often higher than those derived from data \citep{chiappa-etal2024}, so more work on deriving the appropriate modal basis is necessary to make progress on this topic. To this end, our choice of principal components for representing tongue shape features was a pragmatic decision and we do not necessarily think that these represent the empirical correlates of muscle synergies \citep{tresch-etal2006}. Yet we predict that \emph{some} form of empirical modes could represent a kinematic expression of muscle synergies. While statistical modes are clearly a property of the data, there is remarkable uniformity in the low-dimensional patterns of variation across studies. These statistical modes may also be complemented by theoretically-motivated variables, as in guided decompositions using jaw or larynx variables \citep{maeda1990, badin-etal2002}. For example, while our tongue shape modes indirectly encode information on the jaw, with mandibular short tendon as part of the tongue shape representation, they do not allow us to say anything about independent control of the jaw.

\subsection{Limitations and future work}
\label{subsec:limitations}

We have stressed throughout that we aim to address the \emph{sufficiency} of a linear dynamical model for shape dynamics and we cannot conclude that the dynamics of low-dimensional control are truly linear. Analysing the kinematics of a single point on the tongue in these vowels would probably not reveal significant nonlinearity; our previous work shows that a linear harmonic oscillator can be fitted to single key-points in ultrasound data with reasonable accuracy \citep{kirkham-strycharczuk2025}. Indeed, our model is theoretically incapable of capturing nonlinear dependence between movement amplitude and time-to-peak velocity, meaning that alternative mechanisms such as time-varying activation or nonlinear coupling are likely to be necessary.

While not a direct solution to this problem, a natural extension of our model would be to allow coupling between modes. At present, the principal components evolve as uncoupled oscillators, similar to the uncoupled tract variables of task dynamics. This means that the dynamics of one PC are completely independent from the others, with the resulting tongue shape being a simple linear summation via inverted PCA. It is entirely feasible, however, to introduce off-diagonal terms, so that the force acting on one shape mode depends on the position and velocity of the others. The coupled system can then be decomposed into its eigenmodes, with the resulting modes representing a candidate control basis for tongue shape variation that need not coincide with the statistical modes returned by PCA.

Modal coupling is an elegant solution, because the additional temporal complexity is located within the dynamical law itself rather than in a more elaborate control signal. The cost, however, is that the number of parameters increases, which raises a severe identifiability problem for fitting the model to data. This is especially acute for the off-diagonal damping terms, which are expressed only in the velocity of the trajectories, which we were unable to estimate reliably from time-varying PCs. So while such a model cannot be supported by our present empirical data, this extension could motivate the search for alternative dynamical models that can handle a complex modal basis over longer temporal windows.

More practically, our analysis only modelled lax vowels, meaning that highly dynamic movements are not captured here. Previous theoretical work suggests that long vowels and diphthongs in English can be modelled as two separate events \citep{strycharczuk-etal2024}, which would simply represent two model fits, similar to the vowel formation and release in the present study. Furthermore, we used a damped linear point attractor system as an articulatory control law, but we do not claim this is the optimal model form. Indeed, the major downside of modelling short piecewise movements is identifiability, with insufficient information in the signal to disambiguate between different model forms. Finally, our GPA procedure standardizes differences in rotation and displacement. This yields a cross-speaker coordinate system, but has a substantial effect on interpretability, as the modes now represent variation in a shape space, rather than specific constriction degrees. In future work, we plan to analyse the relationship between tongue shape modes and re-derived movement and constriction variables, which will provide a deeper connection between shape modes, task space, and articulator space.

\section{Conclusion}

We advanced a dynamical model of tongue shape dynamics, where low-dimensional empirical variables serve as the control basis for a principled control law. The model sufficiently captures tongue shape dynamics in English monophthongal vowels, with tract and articulatory variables recast as projections from tongue shape space. This reframing also explains how apparent nonlinearities in one-dimensional articulatory measures can arise as artefacts of projection, rather than as properties of the underlying control system.

\section*{Acknowledgments}

This research was funded by The Royal Society grant APX\textbackslash R1\textbackslash 251102 (S.K), which was additionally supported by the Leverhulme Trust, and British Academy Mid-Career Fellowship MFSS24\textbackslash240076 (P.S.). Data collection was originally funded by AHRC grant AH/S011900/1 (P.S., S.K.).

\section*{Author declarations}

The authors have no conflicts of interest to declare. The study received ethical approval from the University of Lancaster Ethics Committee (ref. no. FL18188) and from the University of Manchester’s Proportionate University Research Ethics Committee (ref. no. 2022-13946-22714).


\begin{thebibliography}{65}
\def\enquote#1{``#1,''}
\def\plainquote#1{``#1''}
\expandafter\ifx\csname natexlab\endcsname\relax\def\natexlab#1{#1}\fi
\providecommand{\dourl}[1]{\href{http://#1}{\nolinkurl{#1}}}
\providecommand{\bibinfo}[2]{#2}
\providecommand{\noopsort}[1]{}
\providecommand{\switchargs}[2]{#2#1}
  \def\eatspace #1{#1}

\bibitem[{Badin \emph{et~al.}(2002)Badin, Bailly, Rev{\'e}ret, Baciu,
  Segebarth, and Savariaux}]{badin-etal2002}
\bibinfo{author}{Badin, P.}, \bibinfo{author}{Bailly, G.},
  \bibinfo{author}{Rev{\'e}ret, L.}, \bibinfo{author}{Baciu, M.},
  \bibinfo{author}{Segebarth, C.},  and \bibinfo{author}{Savariaux, C.}
  (\textbf{\bibinfo{year}{2002}}). \enquote{\bibinfo{title}{Three-dimensional
  linear articulatory modeling of tongue, lips and face, based on {MRI} and
  video images}} \bibinfo{journal}{Journal of Phonetics} \textbf{30}(3),
  \bibinfo{pages}{533--553}.

\bibitem[{Beautemps \emph{et~al.}(2001)Beautemps, Badin, and
  Bailly}]{beautemps-etal2001}
\bibinfo{author}{Beautemps, D.}, \bibinfo{author}{Badin, P.},  and
  \bibinfo{author}{Bailly, G.} (\textbf{\bibinfo{year}{2001}}).
  \enquote{\bibinfo{title}{Linear degrees of freedom in speech production:
  Analysis of cineradio- and labio-film data and articulatory-acoustic
  modeling}} \bibinfo{journal}{Journal of the Acoustical Society of America}
  \textbf{109}(5), \bibinfo{pages}{2165--2180}.

\bibitem[{Bernstein(1967)}]{bernstein1967}
\bibinfo{author}{Bernstein, N.} (\textbf{\bibinfo{year}{1967}}).
  \emph{\bibinfo{title}{The Coordination and Regulation of Movements}}
  (\bibinfo{publisher}{Pergamon Press}, \bibinfo{address}{Oxford}).

\bibitem[{Bizzi and Cheung(2013)}]{bizzi-cheung2013}
\bibinfo{author}{Bizzi, E.},  and \bibinfo{author}{Cheung, V. C.~K.}
  (\textbf{\bibinfo{year}{2013}}). \enquote{\bibinfo{title}{The neural origin
  of muscle synergies}} \bibinfo{journal}{Frontiers in Computational
  Neuroscience} \textbf{7}(51), \bibinfo{pages}{1--6}.

\bibitem[{B\"{o}e \emph{et~al.}(1992)B\"{o}e, Perrier, and
  Bailly}]{boe-etal1992}
\bibinfo{author}{B\"{o}e, L.-J.}, \bibinfo{author}{Perrier, P.},  and
  \bibinfo{author}{Bailly, G.} (\textbf{\bibinfo{year}{1992}}).
  \enquote{\bibinfo{title}{The geometric vocal tract variables controlled for
  vowel production: Proposals for constraining acoustic-to-articulatory
  inversion}} \bibinfo{journal}{Journal of Phonetics} \textbf{20}(1),
  \bibinfo{pages}{27--38}.

\bibitem[{Browman and Goldstein(1992)}]{browman-goldstein1992}
\bibinfo{author}{Browman, C.~P.},  and \bibinfo{author}{Goldstein, L.}
  (\textbf{\bibinfo{year}{1992}}). \enquote{\bibinfo{title}{Articulatory
  phonology: An overview}} \bibinfo{journal}{Phonetica} \textbf{49}(3-4),
  \bibinfo{pages}{155--180}.

\bibitem[{Byrd and Saltzman(1998)}]{byrd-saltzman1998}
\bibinfo{author}{Byrd, D.},  and \bibinfo{author}{Saltzman, E.}
  (\textbf{\bibinfo{year}{1998}}). \enquote{\bibinfo{title}{Intragestural
  dynamics of multiple prosodic boundaries}} \bibinfo{journal}{Journal of
  Phonetics} \textbf{26}(2), \bibinfo{pages}{173--199}.

\bibitem[{Calka \emph{et~al.}(2025)Calka, Badin, Nazari, Rochette, Perrier, and
  Payan}]{calka-etal2025}
\bibinfo{author}{Calka, M.}, \bibinfo{author}{Badin, P.},
  \bibinfo{author}{Nazari, M.~A.}, \bibinfo{author}{Rochette, M.},
  \bibinfo{author}{Perrier, P.},  and \bibinfo{author}{Payan, Y.}
  (\textbf{\bibinfo{year}{2025}}). \enquote{\bibinfo{title}{The {G-OBIM} tongue
  model: An accurate open-source biomechanical model of a male human tongue}}
  \bibinfo{journal}{{PLoS} Computational Biology} \textbf{21}(9),
  \bibinfo{pages}{e1013378}.

\bibitem[{Calka \emph{et~al.}(2024)Calka, Perrier, Rochette, and
  Payan}]{calka-etal2024}
\bibinfo{author}{Calka, M.}, \bibinfo{author}{Perrier, P.},
  \bibinfo{author}{Rochette, M.},  and \bibinfo{author}{Payan, Y.}
  (\textbf{\bibinfo{year}{2024}}). \enquote{\bibinfo{title}{Biomechanical
  modeling of the human tongue}} in \emph{\bibinfo{booktitle}{Mechanics of
  Living Tissues: Imaging, Characterization and Modeling Towards the Study of
  Soft Tissues}}, edited by \bibinfo{editor}{C.~Laurent} and
  \bibinfo{editor}{C.~Verdier} (\bibinfo{publisher}{John Wiley \& Sons},
  \bibinfo{address}{Hoboken, NJ}), pp. \bibinfo{pages}{95--115}.

\bibitem[{Chiappa \emph{et~al.}(2024)Chiappa, Tano, Patel, Ingster, Poget, and
  Mathis}]{chiappa-etal2024}
\bibinfo{author}{Chiappa, A.~S.}, \bibinfo{author}{Tano, P.},
  \bibinfo{author}{Patel, N.}, \bibinfo{author}{Ingster, A.},
  \bibinfo{author}{Poget, A.},  and \bibinfo{author}{Mathis, A.}
  (\textbf{\bibinfo{year}{2024}}). \enquote{\bibinfo{title}{Acquiring
  musculoskeletal skills with curriculum-based reinforcement learning}}
  \bibinfo{journal}{Neuron} \textbf{112}(23), \bibinfo{pages}{3969--3983}.

\bibitem[{d'Avella \emph{et~al.}(2003)d'Avella, Saltiel, and
  Bizzi}]{davella-etal2003}
\bibinfo{author}{d'Avella, A.}, \bibinfo{author}{Saltiel, P.},  and
  \bibinfo{author}{Bizzi, E.} (\textbf{\bibinfo{year}{2003}}).
  \enquote{\bibinfo{title}{Combinations of muscle synergies in the construction
  of a natural motor behavior}} \bibinfo{journal}{Nature Neuroscience}
  \textbf{6}(3), \bibinfo{pages}{300--308}.

\bibitem[{Elie \emph{et~al.}(2023)Elie, Lee, and Turk}]{elie-etal2023}
\bibinfo{author}{Elie, B.}, \bibinfo{author}{Lee, D.~N.},  and
  \bibinfo{author}{Turk, A.} (\textbf{\bibinfo{year}{2023}}).
  \enquote{\bibinfo{title}{Modeling trajectories of human speech articulators
  using general {T}au theory}} \bibinfo{journal}{Speech Communication}
  \textbf{151}, \bibinfo{pages}{24--38}.

\bibitem[{Fowler \emph{et~al.}(1980)Fowler, Rubin, Remez, and
  Turvey}]{fowler-etal1980}
\bibinfo{author}{Fowler, C.~A.}, \bibinfo{author}{Rubin, P.},
  \bibinfo{author}{Remez, R.~E.},  and \bibinfo{author}{Turvey, M.~T.}
  (\textbf{\bibinfo{year}{1980}}). \enquote{\bibinfo{title}{Implications for
  speech production of a general theory of action}} in
  \emph{\bibinfo{booktitle}{Language Production}}, edited by
  \bibinfo{editor}{B.~Butterworth} (\bibinfo{publisher}{Academic Press},
  \bibinfo{address}{New York, NY}), pp. \bibinfo{pages}{373--420}.

\bibitem[{Gay \emph{et~al.}(1981)Gay, Lindblom, and Lubker}]{gay-etal1981}
\bibinfo{author}{Gay, T.}, \bibinfo{author}{Lindblom, B.},  and
  \bibinfo{author}{Lubker, J.} (\textbf{\bibinfo{year}{1981}}).
  \enquote{\bibinfo{title}{The production of bite block vowels: Acoustic
  equivalence by selective compensation}} \bibinfo{journal}{Journal of the
  Acoustical Society of America} \textbf{69}(3), \bibinfo{pages}{802--810}.

\bibitem[{Gick(2016)}]{gick2016}
\bibinfo{author}{Gick, B.} (\textbf{\bibinfo{year}{2016}}).
  \enquote{\bibinfo{title}{Ecologizing dimensionality: Prospects for a modular
  theory of speech production}} \bibinfo{journal}{Ecological Psychology}
  \textbf{28}(3), \bibinfo{pages}{176--181}.

\bibitem[{Gick and Stavness(2013)}]{gick-stavness2013}
\bibinfo{author}{Gick, B.},  and \bibinfo{author}{Stavness, I.}
  (\textbf{\bibinfo{year}{2013}}). \enquote{\bibinfo{title}{Modularizing
  speech}} \bibinfo{journal}{Frontiers in Psychology} \textbf{4}(977),
  \bibinfo{pages}{1--3}.

\bibitem[{Gick \emph{et~al.}(2013)Gick, Wilson, and Derrick}]{gick-etal2013}
\bibinfo{author}{Gick, B.}, \bibinfo{author}{Wilson, I.},  and
  \bibinfo{author}{Derrick, D.} (\textbf{\bibinfo{year}{2013}}).
  \emph{\bibinfo{title}{Articulatory Phonetics}}
  (\bibinfo{publisher}{Wiley-Blackwell}, \bibinfo{address}{Chichester}).

\bibitem[{Gubian \emph{et~al.}(2015)Gubian, Torreira, and
  Boves}]{gubian-etal2015}
\bibinfo{author}{Gubian, M.}, \bibinfo{author}{Torreira, F.},  and
  \bibinfo{author}{Boves, L.} (\textbf{\bibinfo{year}{2015}}).
  \enquote{\bibinfo{title}{Using {F}unctional {D}ata {A}nalysis for
  investigating multidimensional dynamic phonetic contrasts}}
  \bibinfo{journal}{Journal of Phonetics} \textbf{49}(1),
  \bibinfo{pages}{16--40}.

\bibitem[{Harshman \emph{et~al.}(1977)Harshman, Ladefoged, and
  Goldstein}]{harshman-etal1977}
\bibinfo{author}{Harshman, R.}, \bibinfo{author}{Ladefoged, P.},  and
  \bibinfo{author}{Goldstein, L.} (\textbf{\bibinfo{year}{1977}}).
  \enquote{\bibinfo{title}{Factor analysis of tongue shapes}}
  \bibinfo{journal}{Journal of the Acoustical Society of America}
  \textbf{62}(3), \bibinfo{pages}{693--707}.

\bibitem[{Honda(1996)}]{honda1996}
\bibinfo{author}{Honda, K.} (\textbf{\bibinfo{year}{1996}}).
  \enquote{\bibinfo{title}{Organization of tongue articulation for vowels}}
  \bibinfo{journal}{Journal of Phonetics} \textbf{24}(1),
  \bibinfo{pages}{39--52}.

\bibitem[{Hoole(1999)}]{hoole1999}
\bibinfo{author}{Hoole, P.} (\textbf{\bibinfo{year}{1999}}).
  \enquote{\bibinfo{title}{On the lingual organization of the {G}erman vowel
  system}} \bibinfo{journal}{Journal of the Acoustical Society of America}
  \textbf{106}(2), \bibinfo{pages}{1020--1032}.

\bibitem[{Iskarous(2005)}]{iskarous2005}
\bibinfo{author}{Iskarous, K.} (\textbf{\bibinfo{year}{2005}}).
  \enquote{\bibinfo{title}{Patterns of tongue movement}}
  \bibinfo{journal}{Journal of Phonetics} \textbf{33},
  \bibinfo{pages}{363--381}.

\bibitem[{Jackson(1988)}]{jackson1988}
\bibinfo{author}{Jackson, M. T.~T.} (\textbf{\bibinfo{year}{1988}}).
  \enquote{\bibinfo{title}{Analysis of tongue positions: Language-specific and
  cross-linguistic models}} \bibinfo{journal}{Journal of the Acoustical Society
  of America} \textbf{84}(1), \bibinfo{pages}{124--143}.

\bibitem[{Kelso \emph{et~al.}(1986)Kelso, Saltzman, and
  Tuller}]{kelso-etal1986}
\bibinfo{author}{Kelso, J.~S.}, \bibinfo{author}{Saltzman, E.~L.},  and
  \bibinfo{author}{Tuller, B.} (\textbf{\bibinfo{year}{1986}}).
  \enquote{\bibinfo{title}{The dynamical perspective on speech production: data
  and theory}} \bibinfo{journal}{Journal of Phonetics} \textbf{14}(1),
  \bibinfo{pages}{29--59}.

\bibitem[{Kent(1992)}]{kent1992}
\bibinfo{author}{Kent, J.~T.} (\textbf{\bibinfo{year}{1992}}).
  \enquote{\bibinfo{title}{New directions in shape analysis}} in
  \emph{\bibinfo{booktitle}{The Art of Statistical Science}}, edited by
  \bibinfo{editor}{K.~V. Mardia} (\bibinfo{publisher}{Wiley},
  \bibinfo{address}{Chichester}), pp. \bibinfo{pages}{115--127}.

\bibitem[{Kirkham(2025{\natexlab{a}})}]{kirkham2025b}
\bibinfo{author}{Kirkham, S.} (\textbf{\bibinfo{year}{2025}}{\natexlab{a}}).
  \enquote{\bibinfo{title}{Discovering dynamical laws for speech gestures}}
  \bibinfo{journal}{Cognitive Science} \textbf{49}(5), \bibinfo{pages}{e70064}.

\bibitem[{Kirkham(2025{\natexlab{b}})}]{kirkham2025}
\bibinfo{author}{Kirkham, S.} (\textbf{\bibinfo{year}{2025}}{\natexlab{b}}).
  \enquote{\bibinfo{title}{Scaling laws for nonlinear dynamical models of
  articulatory control}} \bibinfo{journal}{{JASA} Express Letters}
  \textbf{5}(2), \bibinfo{pages}{1--7}.

\bibitem[{Kirkham and Strycharczuk(2025)}]{kirkham-strycharczuk2025}
\bibinfo{author}{Kirkham, S.},  and \bibinfo{author}{Strycharczuk, P.}
  (\textbf{\bibinfo{year}{2025}}). \enquote{\bibinfo{title}{Dynamical model
  parameters from ultrasound tongue kinematics}} \bibinfo{journal}{JASA Express
  Letters} .

\bibitem[{Lloyd \emph{et~al.}(2012)Lloyd, Stavness, and Fels}]{lloyd-etal2012}
\bibinfo{author}{Lloyd, J.~E.}, \bibinfo{author}{Stavness, I.},  and
  \bibinfo{author}{Fels, S.} (\textbf{\bibinfo{year}{2012}}).
  \enquote{\bibinfo{title}{Artisynth: A fast interactive biomechanical modeling
  toolkit combining multibody and finite element simulation}} in
  \emph{\bibinfo{booktitle}{Soft Tissue Biomechanical Modeling for Computer
  Assisted Surgery}}, edited by \bibinfo{editor}{Y.~Payan}
  (\bibinfo{publisher}{Springer}, \bibinfo{address}{Berlin}), pp.
  \bibinfo{pages}{355--394}.

\bibitem[{Lo \emph{et~al.}(2025)Lo, Strycharczuk, and Kirkham}]{lo-etal2025}
\bibinfo{author}{Lo, J. J.~H.}, \bibinfo{author}{Strycharczuk, P.},  and
  \bibinfo{author}{Kirkham, S.} (\textbf{\bibinfo{year}{2025}}).
  \enquote{\bibinfo{title}{Articulatory strategy in vowel production as a basis
  for speaker discrimination}} \bibinfo{journal}{Proceedings of Interspeech
  2025} \bibinfo{pages}{3504--3508}.

\bibitem[{Maeda(1990)}]{maeda1990}
\bibinfo{author}{Maeda, S.} (\textbf{\bibinfo{year}{1990}}).
  \enquote{\bibinfo{title}{Compensatory articulation during speech: Evidence
  for the analysis and synthesis of vocal-tract shapes using an articulatory
  model}} in \emph{\bibinfo{booktitle}{Speech Production and Speech
  Modelling}}, edited by \bibinfo{editor}{W.~J. Hardcastle} and
  \bibinfo{editor}{A.~Marchal} (\bibinfo{publisher}{Springer},
  \bibinfo{address}{Dordrecht}), pp. \bibinfo{pages}{131--149}.

\bibitem[{Maeda(1991)}]{maeda1991}
\bibinfo{author}{Maeda, S.} (\textbf{\bibinfo{year}{1991}}).
  \enquote{\bibinfo{title}{On articulatory and acoustic variabilities}}
  \bibinfo{journal}{Journal of Phonetics} \textbf{19}(3--4),
  \bibinfo{pages}{321--331}.

\bibitem[{Maeda and Honda(1994)}]{maeda-honda1994}
\bibinfo{author}{Maeda, S.},  and \bibinfo{author}{Honda, K.}
  (\textbf{\bibinfo{year}{1994}}). \enquote{\bibinfo{title}{From {EMG} to
  formant patterns of vowels: The implication of vowel spaces}}
  \bibinfo{journal}{Phonetica} \textbf{51}(1--3), \bibinfo{pages}{17--29}.

\bibitem[{Mathis \emph{et~al.}(2018)Mathis, Mamidanna, Mury, Abe, Murthy,
  Mathis, and Bethge}]{mathis-etal2018}
\bibinfo{author}{Mathis, A.}, \bibinfo{author}{Mamidanna, P.},
  \bibinfo{author}{Mury, K.~M.}, \bibinfo{author}{Abe, T.},
  \bibinfo{author}{Murthy, V.~N.}, \bibinfo{author}{Mathis, M.~W.},  and
  \bibinfo{author}{Bethge, M.} (\textbf{\bibinfo{year}{2018}}).
  \enquote{\bibinfo{title}{{DeepLabCut}: Markerless pose estimation of
  user-defined body parts with deep learning}} \bibinfo{journal}{Nature
  Neuroscience} \textbf{21}(9), \bibinfo{pages}{1281--1289}.

\bibitem[{Mori(1965)}]{mori1965}
\bibinfo{author}{Mori, H.} (\textbf{\bibinfo{year}{1965}}).
  \enquote{\bibinfo{title}{Transport, collective motion, and {B}rownian
  motion}} \bibinfo{journal}{Progress of Theoretical Physics} \textbf{33}(3),
  \bibinfo{pages}{423--455}.

\bibitem[{Mussa-Ivaldi \emph{et~al.}(1994)Mussa-Ivaldi, Giszter, and
  Bizzi}]{mussa-ivaldi-etal1994}
\bibinfo{author}{Mussa-Ivaldi, F.}, \bibinfo{author}{Giszter, S.},  and
  \bibinfo{author}{Bizzi, E.} (\textbf{\bibinfo{year}{1994}}).
  \enquote{\bibinfo{title}{Linear combinations of primitives in vertebrate
  motor control}} \bibinfo{journal}{Proceedings of the National Academy of
  Sciences} \textbf{91}(16), \bibinfo{pages}{7534--7538}.

\bibitem[{Nix \emph{et~al.}(1996)Nix, Pap\c{c}un, Hogden, and
  Zlokarnik}]{nix-etal1996}
\bibinfo{author}{Nix, D.~A.}, \bibinfo{author}{Pap\c{c}un, G.},
  \bibinfo{author}{Hogden, J.},  and \bibinfo{author}{Zlokarnik, I.}
  (\textbf{\bibinfo{year}{1996}}). \enquote{\bibinfo{title}{Two
  cross-linguistic factors underlying tongue shapes for vowels}}
  \bibinfo{journal}{Journal of the Acoustical Society of America}
  \textbf{99}(6), \bibinfo{pages}{3707--3717}.

\bibitem[{Ostry and Munhall(1985)}]{ostry-munhall1985}
\bibinfo{author}{Ostry, D.~J.},  and \bibinfo{author}{Munhall, K.~G.}
  (\textbf{\bibinfo{year}{1985}}). \enquote{\bibinfo{title}{Control of rate and
  duration of speech movements}} \bibinfo{journal}{Journal of the Acoustical
  Society of America} \textbf{77}(2), \bibinfo{pages}{640--648}.

\bibitem[{Payan and Perrier(1997)}]{payan-perrier1997}
\bibinfo{author}{Payan, Y.},  and \bibinfo{author}{Perrier, P.}
  (\textbf{\bibinfo{year}{1997}}). \enquote{\bibinfo{title}{Synthesis of {V-V}
  sequences with a {2D} biomechanical tongue model controlled by the
  {E}quilibrium {P}oint {H}ypothesis}} \bibinfo{journal}{Speech Communication}
  \textbf{22}(2--3), \bibinfo{pages}{185--205}.

\bibitem[{Perrier \emph{et~al.}(1996)Perrier, L{\oe}venbruck, and
  Payan}]{perrier-etal1996}
\bibinfo{author}{Perrier, P.}, \bibinfo{author}{L{\oe}venbruck, H.},  and
  \bibinfo{author}{Payan, Y.} (\textbf{\bibinfo{year}{1996}}).
  \enquote{\bibinfo{title}{Control of tongue movements in speech: The
  {E}quilibrium {P}oint {H}ypothesis perspective}} \bibinfo{journal}{Journal of
  Phonetics} \textbf{24}(1), \bibinfo{pages}{53--75}.

\bibitem[{Ramanarayanan \emph{et~al.}(2013)Ramanarayanan, Goldstein, and
  Narayanan}]{ramanarayanan-etal2013}
\bibinfo{author}{Ramanarayanan, V.}, \bibinfo{author}{Goldstein, L.},  and
  \bibinfo{author}{Narayanan, S.~S.} (\textbf{\bibinfo{year}{2013}}).
  \enquote{\bibinfo{title}{Spatio-temporal articulatory movement primitives
  during speech production: Extraction, interpretation, and validation}}
  \bibinfo{journal}{Journal of the Acoustical Society of America}
  \textbf{134}(2), \bibinfo{pages}{1378--1394}.

\bibitem[{Saltzman and Munhall(1989)}]{saltzman-munhall1989}
\bibinfo{author}{Saltzman, E.},  and \bibinfo{author}{Munhall, K.~G.}
  (\textbf{\bibinfo{year}{1989}}). \enquote{\bibinfo{title}{A dynamical
  approach to gestural patterning in speech production}}
  \bibinfo{journal}{Ecological Psychology} \textbf{1}(4),
  \bibinfo{pages}{333--382}.

\bibitem[{Serrurier \emph{et~al.}(2019)Serrurier, Badin, Lamalle, and
  Neuschaefer-Rube}]{serrurier-etal2019}
\bibinfo{author}{Serrurier, A.}, \bibinfo{author}{Badin, P.},
  \bibinfo{author}{Lamalle, L.},  and \bibinfo{author}{Neuschaefer-Rube, C.}
  (\textbf{\bibinfo{year}{2019}}). \enquote{\bibinfo{title}{Characterization of
  inter-speaker articulatory variability: A two-level multi-speaker modelling
  approach based on {MRI} data}} \bibinfo{journal}{Journal of the Acoustical
  Society of America} \textbf{145}(4), \bibinfo{pages}{2149--2170}.

\bibitem[{Slud \emph{et~al.}(2002)Slud, Stone, Smith, and
  Goldstein}]{slud-etal2002}
\bibinfo{author}{Slud, E.}, \bibinfo{author}{Stone, M.},
  \bibinfo{author}{Smith, P.~J.},  and \bibinfo{author}{Goldstein, M.~H.}
  (\textbf{\bibinfo{year}{2002}}). \enquote{\bibinfo{title}{Principal
  components representation of the two-dimensional coronal tongue surface}}
  \bibinfo{journal}{Phonetica} \textbf{59}(2--3), \bibinfo{pages}{108--133}.

\bibitem[{Sorensen and Gafos(2016)}]{sorensen-gafos2016}
\bibinfo{author}{Sorensen, T.},  and \bibinfo{author}{Gafos, A.~I.}
  (\textbf{\bibinfo{year}{2016}}). \enquote{\bibinfo{title}{The gesture as an
  autonomous nonlinear dynamical system}} \bibinfo{journal}{Ecological
  Psychology} \textbf{28}(4), \bibinfo{pages}{188--215}.

\bibitem[{Sorensen \emph{et~al.}(2019)Sorensen, Toutios, Goldstein, and
  Narayanan}]{sorensen-etal2019}
\bibinfo{author}{Sorensen, T.}, \bibinfo{author}{Toutios, A.},
  \bibinfo{author}{Goldstein, L.},  and \bibinfo{author}{Narayanan, S.}
  (\textbf{\bibinfo{year}{2019}}). \enquote{\bibinfo{title}{Task-dependence of
  articulator synergies}} \bibinfo{journal}{Journal of the Acoustical Society
  of America} \textbf{145}(3), \bibinfo{pages}{1504--1520}.

\bibitem[{Stern and Shaw(2025)}]{stern-shaw2025}
\bibinfo{author}{Stern, M.~C.},  and \bibinfo{author}{Shaw, J.~A.}
  (\textbf{\bibinfo{year}{2025}}). \enquote{\bibinfo{title}{Nonlinear
  second-order dynamics describe labial constriction trajectories across
  languages and contexts}} \bibinfo{journal}{Journal of Phonetics}
  \textbf{111}(101427), \bibinfo{pages}{1--18}.

\bibitem[{Stone \emph{et~al.}(1997)Stone, Goldstein, and
  Zhang}]{stone-etal1997}
\bibinfo{author}{Stone, M.}, \bibinfo{author}{Goldstein, M.~H.},  and
  \bibinfo{author}{Zhang, Y.} (\textbf{\bibinfo{year}{1997}}).
  \enquote{\bibinfo{title}{Principal component analysis of cross sections of
  tongue shapes in vowel production}} \bibinfo{journal}{Speech Communication}
  \textbf{22}(2--3), \bibinfo{pages}{173--184}.

\bibitem[{Stone \emph{et~al.}(2018)Stone, Woo, Lee, Poole, Seagraves, Chung,
  Kim, Murano, Prince, and Blemker}]{stone-etal2018}
\bibinfo{author}{Stone, M.}, \bibinfo{author}{Woo, J.}, \bibinfo{author}{Lee,
  J.}, \bibinfo{author}{Poole, T.}, \bibinfo{author}{Seagraves, A.},
  \bibinfo{author}{Chung, M.}, \bibinfo{author}{Kim, E.},
  \bibinfo{author}{Murano, E.~Z.}, \bibinfo{author}{Prince, J.~L.},  and
  \bibinfo{author}{Blemker, S.~S.} (\textbf{\bibinfo{year}{2018}}).
  \enquote{\bibinfo{title}{Structure and variability in human tongue muscle
  anatomy}} \bibinfo{journal}{Computer Methods in Biomechanics and Biomedical
  Engineering: Imaging \& Visualization} \textbf{6}(5),
  \bibinfo{pages}{499--507}.

\bibitem[{Story(2005)}]{story2005}
\bibinfo{author}{Story, B.~H.} (\textbf{\bibinfo{year}{2005}}).
  \enquote{\bibinfo{title}{Synergistic modes of vocal tract articulation for
  {A}merican {E}nglish vowels}} \bibinfo{journal}{Journal of the Acoustical
  Society of America} \textbf{118}(6), \bibinfo{pages}{3834--3859}.

\bibitem[{Story(2007)}]{story2007}
\bibinfo{author}{Story, B.~H.} (\textbf{\bibinfo{year}{2007}}).
  \enquote{\bibinfo{title}{Time dependence of vocal tract modes during
  production of vowels and vowel sequences}} \bibinfo{journal}{Journal of the
  Acoustical Society of America} \textbf{121}(6), \bibinfo{pages}{3770--3789}.

\bibitem[{Story and Titze(1998)}]{story-titze1998}
\bibinfo{author}{Story, B.~H.},  and \bibinfo{author}{Titze, I.~R.}
  (\textbf{\bibinfo{year}{1998}}). \enquote{\bibinfo{title}{Parameterization of
  vocal tract area functions by empirical orthogonal modes}}
  \bibinfo{journal}{Journal of Phonetics} \textbf{26}(3),
  \bibinfo{pages}{223--260}.

\bibitem[{Strycharczuk and Kirkham(2025)}]{strycharczuk-kirkham2025}
\bibinfo{author}{Strycharczuk, P.},  and \bibinfo{author}{Kirkham, S.}
  (\textbf{\bibinfo{year}{2025}}). \enquote{\bibinfo{title}{Articulatory
  strategies in male and female vowel production}} \bibinfo{journal}{Journal of
  Speech, Language, and Hearing Research} \textbf{68}(12),
  \bibinfo{pages}{5629--5649}.

\bibitem[{Strycharczuk \emph{et~al.}(2024)Strycharczuk, Kirkham, Gorman, and
  Nagamine}]{strycharczuk-etal2024}
\bibinfo{author}{Strycharczuk, P.}, \bibinfo{author}{Kirkham, S.},
  \bibinfo{author}{Gorman, E.},  and \bibinfo{author}{Nagamine, T.}
  (\textbf{\bibinfo{year}{2024}}). \enquote{\bibinfo{title}{Towards a dynamical
  model of {E}nglish vowels: Evidence from diphthongisation}}
  \bibinfo{journal}{Journal of Phonetics} \textbf{107}, \bibinfo{pages}{1--26}.

\bibitem[{Strycharczuk \emph{et~al.}(2025)Strycharczuk, Kirkham, Gorman, and
  Nagamine}]{strycharczuk-etal2025}
\bibinfo{author}{Strycharczuk, P.}, \bibinfo{author}{Kirkham, S.},
  \bibinfo{author}{Gorman, E.},  and \bibinfo{author}{Nagamine, T.}
  (\textbf{\bibinfo{year}{2025}}). \enquote{\bibinfo{title}{Dimensionality
  reduction in lingual articulation of vowels: Evidence from lax vowels in
  {N}orthern {A}nglo-{E}nglish}} \bibinfo{journal}{Language and Speech}
  \textbf{68}(3), \bibinfo{pages}{689--721}.

\bibitem[{Strycharczuk \emph{et~al.}(2026)Strycharczuk, Lo, and
  Kirkham}]{strycharczuk-etal2026}
\bibinfo{author}{Strycharczuk, P.}, \bibinfo{author}{Lo, J. J.~H.},  and
  \bibinfo{author}{Kirkham, S.} (\textbf{\bibinfo{year}{2026}}).
  \enquote{\bibinfo{title}{Articulatory strategy as a source of variation in
  acoustic vowel dynamics}} \bibinfo{journal}{Journal of the Acoustical Society
  of America} \textbf{159}(5), \bibinfo{pages}{4068--4078}.

\bibitem[{Strycharczuk \emph{et~al.}(2020)Strycharczuk,
  L\'{o}pez-Ib\'{a}\~{n}ez, Brown, and Leemann}]{strycharczuk-etal2020b}
\bibinfo{author}{Strycharczuk, P.}, \bibinfo{author}{L\'{o}pez-Ib\'{a}\~{n}ez,
  M.}, \bibinfo{author}{Brown, G.},  and \bibinfo{author}{Leemann, A.}
  (\textbf{\bibinfo{year}{2020}}). \enquote{\bibinfo{title}{{G}eneral
  {N}orthern {E}nglish: Exploring regional variation in the north of {E}ngland
  with machine learning}} \bibinfo{journal}{Frontiers in Artificial
  Intelligence} \textbf{3}(48), \bibinfo{pages}{1--18}.

\bibitem[{Takano and Honda(2007)}]{takano-honda2007}
\bibinfo{author}{Takano, S.},  and \bibinfo{author}{Honda, K.}
  (\textbf{\bibinfo{year}{2007}}). \enquote{\bibinfo{title}{An {MRI} analysis
  of the extrinsic tongue muscles during vowel production}}
  \bibinfo{journal}{Speech Communication} \textbf{49}(1),
  \bibinfo{pages}{49--58}.

\bibitem[{Tilsen(2020)}]{tilsen2020}
\bibinfo{author}{Tilsen, S.} (\textbf{\bibinfo{year}{2020}}).
  \enquote{\bibinfo{title}{A different view of gestural activation: Learning
  gestural parameters and activations with an {RNN}}} \bibinfo{journal}{Cornell
  Working Papers in Phonetics and Phonology} \bibinfo{pages}{1--49}.

\bibitem[{Tresch \emph{et~al.}(2006)Tresch, Cheung, and
  d'Avella}]{tresch-etal2006}
\bibinfo{author}{Tresch, M.~C.}, \bibinfo{author}{Cheung, V. C.~K.},  and
  \bibinfo{author}{d'Avella, A.} (\textbf{\bibinfo{year}{2006}}).
  \enquote{\bibinfo{title}{Matrix factorization algorithms for the
  identification of muscle synergies: Evaluation on simulated and experimental
  data sets}} \bibinfo{journal}{Journal of Neurophysiology} \textbf{95}(4),
  \bibinfo{pages}{2199--2212}.

\bibitem[{Woo \emph{et~al.}(2021)Woo, Xing, Prince, Stone, Gomez, Reese,
  Wedeen, and El~Fakhri}]{woo-etal2021}
\bibinfo{author}{Woo, J.}, \bibinfo{author}{Xing, F.}, \bibinfo{author}{Prince,
  J.~L.}, \bibinfo{author}{Stone, M.}, \bibinfo{author}{Gomez, A.~D.},
  \bibinfo{author}{Reese, T.~G.}, \bibinfo{author}{Wedeen, V.~J.},  and
  \bibinfo{author}{El~Fakhri, G.} (\textbf{\bibinfo{year}{2021}}).
  \enquote{\bibinfo{title}{A deep joint sparse non-negative matrix
  factorization framework for identifying the common and subject-specific
  functional units of tongue motion during speech}} \bibinfo{journal}{Medical
  Image Analysis} \textbf{72}, \bibinfo{pages}{102131}.

\bibitem[{Wrench and Balch-Tomes(2022)}]{wrench-balch-tomes2022}
\bibinfo{author}{Wrench, A.},  and \bibinfo{author}{Balch-Tomes, J.}
  (\textbf{\bibinfo{year}{2022}}). \enquote{\bibinfo{title}{Beyond the edge:
  Markerless pose estimation of speech articulators from ultrasound and camera
  images using {DeepLabCut}}} \bibinfo{journal}{Sensors} \textbf{22}(1133),
  \bibinfo{pages}{1--29}.

\bibitem[{Wrench(2024)}]{wrench2024}
\bibinfo{author}{Wrench, A.~A.} (\textbf{\bibinfo{year}{2024}}).
  \enquote{\bibinfo{title}{The compartmental tongue}} \bibinfo{journal}{Journal
  of Speech, Language, and Hearing Research} \textbf{67}(10S),
  \bibinfo{pages}{3887--3913}.

\bibitem[{Zheng \emph{et~al.}(2003)Zheng, Hasegawa-Johnson, and
  Pizza}]{zheng-etal2003}
\bibinfo{author}{Zheng, Y.}, \bibinfo{author}{Hasegawa-Johnson, M.},  and
  \bibinfo{author}{Pizza, S.} (\textbf{\bibinfo{year}{2003}}).
  \enquote{\bibinfo{title}{Analysis of the three-dimensional tongue shape using
  a three-index factor analysis model}} \bibinfo{journal}{Journal of the
  Acoustical Society of America} \textbf{113}(1), \bibinfo{pages}{478--486}.

\bibitem[{Zwanzig(1961)}]{zwanzig1961}
\bibinfo{author}{Zwanzig, R.} (\textbf{\bibinfo{year}{1961}}).
  \enquote{\bibinfo{title}{Memory effects in irreversible thermodynamics}}
  \bibinfo{journal}{Physical Review} \textbf{124}(4),
  \bibinfo{pages}{983--992}.

\end{thebibliography}
\end{document}